\documentclass[10pt,twocolumn,letterpaper]{article}

\usepackage[pagenumbers]{cvpr}

\usepackage{pifont}   
\usepackage{placeins} 
\usepackage{caption}  
\usepackage{multirow}
\usepackage[percent]{overpic}

\definecolor{cvprblue}{rgb}{0.21,0.49,0.74}
\usepackage[pagebackref,breaklinks,colorlinks,allcolors=cvprblue]{hyperref}
\graphicspath{{figures/}}

\newcommand{\qrow}[2]{%
  \rotatebox{90}{\parbox{0.08\textwidth}{\centering\scriptsize #1}}&%
  \includegraphics[width=0.188\textwidth]{panovideo/longrange_s4_0148_840066/#2_f00}&%
  \begin{overpic}[width=0.188\textwidth]{panovideo/longrange_s4_0148_840066/#2_f20}%
    \put(70,2){\color{red}\linethickness{1pt}\framebox(28,15){}}%
  \end{overpic}&%
  \begin{overpic}[width=0.188\textwidth]{panovideo/longrange_s4_0148_840066/#2_f40}%
    \put(50,5){\color{red}\linethickness{1pt}\framebox(25,15){}}%
  \end{overpic}&%
  \begin{overpic}[width=0.188\textwidth]{panovideo/longrange_s4_0148_840066/#2_f60}%
    \put(10,2){\color{red}\linethickness{1pt}\framebox(50,20){}}%
  \end{overpic}&%
  \begin{overpic}[width=0.188\textwidth]{panovideo/longrange_s4_0148_840066/#2_f80}%
    \put(30,5){\color{red}\linethickness{1pt}\framebox(25,15){}}%
  \end{overpic}\\}

\newcommand{\qrowGBoxLastThree}[3]{%
  \rotatebox{90}{\parbox{0.08\textwidth}{\centering\scriptsize #1}}&%
  \includegraphics[width=0.188\textwidth]{#2/#3_f00}&%
  \includegraphics[width=0.188\textwidth]{#2/#3_f20}&%
  \begin{overpic}[width=0.188\textwidth]{#2/#3_f40}%
    \put(35,10){\color{red}\linethickness{1pt}\framebox(40,30){}}%
  \end{overpic}&%
  \begin{overpic}[width=0.188\textwidth]{#2/#3_f60}%
    \put(35,10){\color{red}\linethickness{1pt}\framebox(40,30){}}%
  \end{overpic}&%
  \begin{overpic}[width=0.188\textwidth]{#2/#3_f80}%
    \put(35,10){\color{red}\linethickness{1pt}\framebox(40,30){}}%
  \end{overpic}\\}

\newcommand{\qrowGBoxfddgs}[2]{%
  \rotatebox{90}{\parbox{0.08\textwidth}{\centering\scriptsize #1}}&%
  \includegraphics[width=0.188\textwidth]{#2_00}&%
  \includegraphics[width=0.188\textwidth]{#2_01}&%
  \includegraphics[width=0.188\textwidth]{#2_02}&%
  \includegraphics[width=0.188\textwidth]{#2_03}&%
  \includegraphics[width=0.188\textwidth]{#2_04}\\}

\newcommand{\qrowGBoxFourth}[3]{%
  \rotatebox{90}{\parbox{0.08\textwidth}{\centering\scriptsize #1}}&%
  \includegraphics[width=0.188\textwidth]{#2/#3_f00}&%
  \includegraphics[width=0.188\textwidth]{#2/#3_f20}&%
  \includegraphics[width=0.188\textwidth]{#2/#3_f40}&%
  \begin{overpic}[width=0.188\textwidth]{#2/#3_f60}%
    \put(10,5){\color{red}\linethickness{1pt}\framebox(50,25){}}%
  \end{overpic}&%
  \includegraphics[width=0.188\textwidth]{#2/#3_f80}\\}

\newcommand{\qrowGBoxMiddleThree}[3]{%
  \rotatebox{90}{\parbox{0.08\textwidth}{\centering\scriptsize #1}}&%
  \includegraphics[width=0.188\textwidth]{#2/#3_f00}&%
  \begin{overpic}[width=0.188\textwidth]{#2/#3_f20}%
    \put(50,22.5){\color{red}\linethickness{1pt}\framebox(20,12.5){}}%
  \end{overpic}&%
  \begin{overpic}[width=0.188\textwidth]{#2/#3_f40}%
    \put(65,25){\color{red}\linethickness{1pt}\framebox(10,7.5){}}%
  \end{overpic}&%
  \begin{overpic}[width=0.188\textwidth]{#2/#3_f60}%
    \put(65,22.5){\color{red}\linethickness{1pt}\framebox(15,17.5){}}%
  \end{overpic}&%
  \includegraphics[width=0.188\textwidth]{#2/#3_f80}\\}

\title{Genie Sim PanoWorld: An Infinite Indoor 3D World Generation Pipeline via Panoramic Scene Modeling and Simulation}
\author{Yongxin Su$^{1}$ \and
        Linjie Hou$^{1}$ \and
        Feng Wang$^{1}$ \and
        Jialin Tang$^{1}$ \and
        Zhijun Li$^{2}$ \and
        Qian Wang$^{\dagger}$ \and
        Maoqing Yao$^{\dagger}$}
\date{}

\begin{document}
\makeatletter
\twocolumn[{%
  \@maketitle
  \vspace*{-8pt}
  \begin{minipage}[t]{0.48\textwidth}
    \vspace*{-280pt}
    \begin{abstract}
      We address the problem of reconstructing a high-fidelity, freely navigable 3D scene from a single
    $360^\circ$ panorama, without per-scene optimization or multi-view capture.
    Existing methods either lack metric trajectory control, which hinders reliable downstream
    3D reconstruction, or struggle with large disocclusions under long-range camera motion
    while requiring high-end multi-GPU servers.
    We present \textbf{Genie Sim PanoWorld}, a two-stage feed-forward pipeline that bridges generation
    and reconstruction via an explicit, trajectory-controllable panoramic video. A NavMesh-planned $\mathrm{SE}(3)$ roaming trajectory is injected into a latent
    video diffusion model through dense geometry-warped conditioning; long--short trajectory
    mixed training and a self-consistency objective based on shortcut models together yield high-fidelity
    video in four CFG-free denoising steps.
    A feed-forward panoramic reconstructor then lifts the generated video into a high-fidelity 
    3D Gaussian scene that supports real-time, free-viewpoint roaming and can be directly used as a simulation-ready asset for embodied AI applications. Experiments show that \textbf{Genie Sim PanoWorld} 
    outperforms geometry-conditioned baselines in both panoramic video generation and 
    downstream 3D reconstruction, while generalizing zero-shot to unseen indoor scenes.
    \end{abstract}
  \end{minipage}
  \hfill
  \begin{minipage}[t]{0.48\textwidth}
    \centering
    \includegraphics[
      width=\textwidth,
      clip
    ]{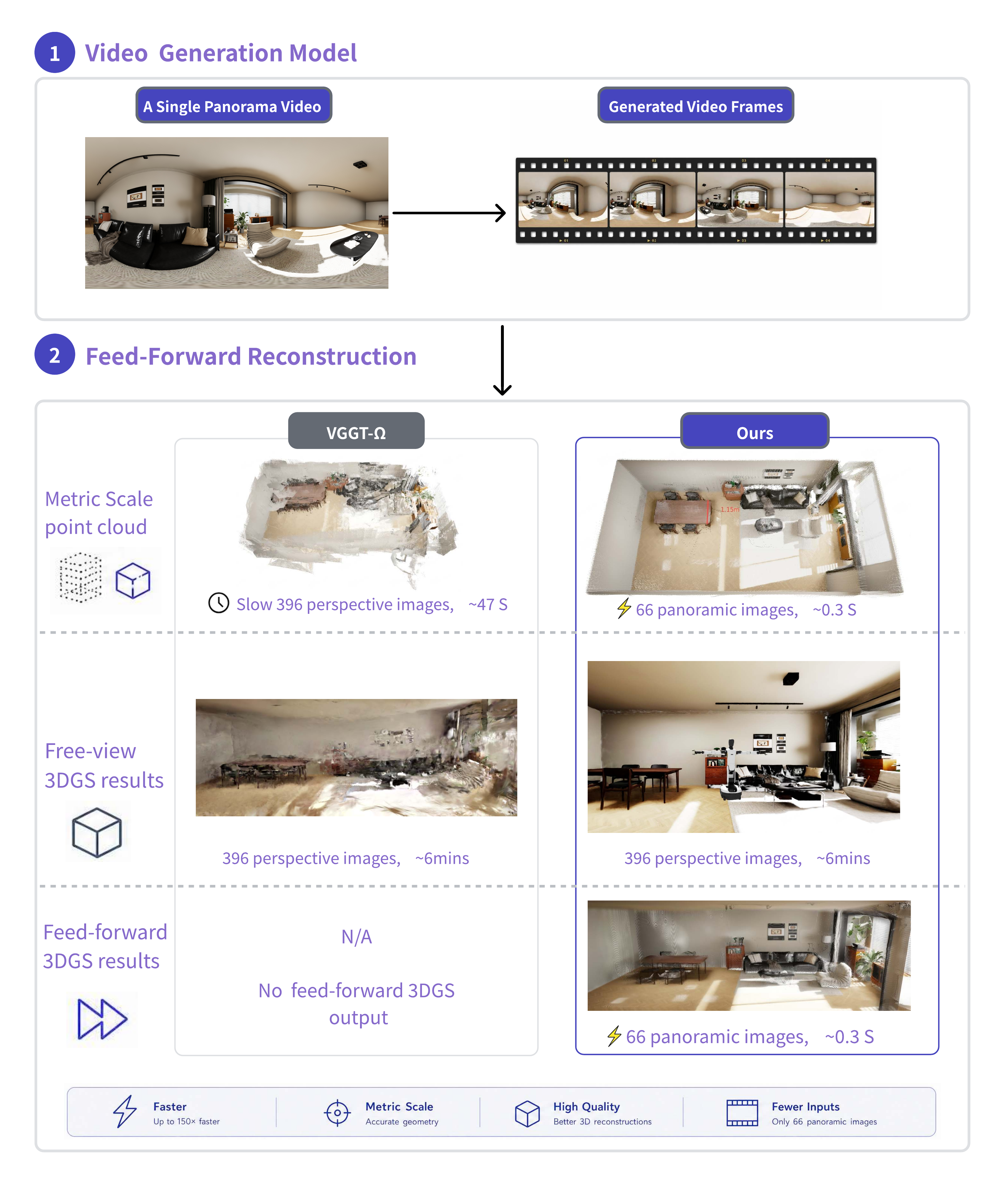}
    \captionof{figure}{
      We propose Genie Sim PanoWorld, which requires only an indoor panoramic
      image and can directly produce a metric-scale 3D point cloud, accurate
      poses, and feed-forward Gaussian results within seconds. It can also be
      provided directly to the downstream per-scene-optimized 3DGS to generate
      embodied intelligence that can use high-fidelity digital twin assets.
    }
    \label{fig:main}
  \end{minipage}
  \vspace{6pt}
}]
\makeatother

\renewcommand{\thefootnote}{\fnsymbol{footnote}}
\footnotetext[2] {Corresponding author.}
\renewcommand{\thefootnote}{\arabic{footnote}}

\vspace*{-60pt}
\section{Introduction}
Creating immersive, explorable 3D indoor environments has long been a goal in vision and
graphics, with applications in virtual tours, embodied AI, and content creation. The
classical route to such a scene is per-scene reconstruction: given many calibrated views,
whose poses are typically recovered by structure-from-motion, optimize a radiance field such
as NeRF~\cite{mildenhall2020nerf} or 3D Gaussian Splatting~\cite{kerbl20233dgs} to fit the
captured views. These methods produce high-quality free-viewpoint renderings, but they
require many overlapping views of the scene and a lengthy per-scene optimization, and
they only interpolate among the observed regions rather than synthesizing unseen content.

A single $360^\circ$ panorama is an appealing starting point that sidesteps the
multi-view capture burden: it captures a whole room in one shot, is cheap to acquire, and is
already the common medium for real-estate and VR-tour products. Yet a panorama is observed
from one fixed viewpoint. It records the scene only as seen from that single location, so a
user can look around freely but cannot move far because the input provides neither parallax
cues nor observations of content hidden behind occluders. Such dense multi-view coverage is
precisely what per-scene reconstruction relies on. Turning such a single panorama into a freely roamable, full $6$-DoF scene in a
feed-forward manner, without any per-scene optimization, is the central challenge: the
single view must be turned into consistent multi-view evidence and lifted into a renderable
3D model in one forward pass.

Existing approaches fall short of this goal in different ways. One line of work generates
panoramic imagery, videos, or multi-view observations through free-form or text-guided
sampling rather than along a prescribed camera trajectory~\cite{NEURIPS2025_d9a12369,yinPanoWorldXGeneratingExplorable2025}.
Without explicit metric trajectory and geometric camera control, the generated views cannot
be reliably reconstructed into a roamable 3D model.
A second line couples camera-guided generation with explicit
geometry~\cite{yangMatrix3DOmnidirectionalExplorable2025,Zhang_2026_CVPR,liuOmniRoamWorldWandering2026},
but tends to degrade under long indoor camera displacements, where the geometric warp
becomes unreliable and disocclusions
dominate~\cite{yangMatrix3DOmnidirectionalExplorable2025}, or discards the metric geometry
needed to follow a collision-free path~\cite{liuOmniRoamWorldWandering2026}. Many of these
systems also build on large backbones with many-step, classifier-free-guided sampling, so
they require high-end multi-GPU
servers~\cite{shen2026lyra20explorablegenerative,hy-worldHYWorld20MultiModal2026} and are
hard to deploy on commodity hardware.

We present \textbf{Genie Sim PanoWorld}, a feed-forward pipeline that turns a single ERP panorama
into a roamable 3D Gaussian scene. The pipeline bridges generation and reconstruction with
an explicit, trajectory-controllable panoramic video. From the input panorama we
recover a coarse scene mesh, plan a collision-free roaming trajectory on a navigation mesh,
and synthesize a panoramic video that follows that path while inpainting newly disoccluded
regions. The generated video provides dense multi-view observations of the scene; a
feed-forward panoramic reconstructor then estimates per-frame camera poses and geometry, and
a Gaussian-splatting decoder lifts them into a 3D Gaussian model for real-time free-viewpoint
rendering. The two stages support each other: the trajectory-controllable video supplies
wide-baseline but geometrically consistent views that make single-panorama reconstruction
tractable, while the metric trajectory keeps both stages in one coherent frame. Two
components keep generation practical. A \emph{long--short trajectory mixed training} scheme
couples high-fidelity short-range supervision with large-disocclusion long-range supervision,
and a \emph{self-consistency objective based on shortcut models}~\cite{frans2025oneshortcut} eliminates the need for classifier-free guidance and reduces
sampling to four steps. As a result, the full single-panorama-to-scene pipeline runs without
any test-time optimization on a single consumer GPU.

Our contributions are as follows.
\begin{itemize}
  \item A feed-forward single-panorama-to-3D pipeline that couples
        trajectory-controllable panoramic video generation with Gaussian
        reconstruction, enabling free-viewpoint rendering without per-scene
        optimization.
  \item A NavMesh-guided panoramic video generator with metric $\mathrm{SE}(3)$
        trajectory control, long--short trajectory mixed training, and self-consistency
        training for shortcut models, achieving improved fidelity and trajectory accuracy
        while substantially reducing inference cost.
  \item A feed-forward panoramic reconstructor with a voxel-aligned Gaussian
        decoder that recovers metric poses, geometry, and renderable 3D Gaussian
        primitives from wide-baseline panoramic observations.
  \item A large-scale indoor panoramic dataset providing dense multi-view
        trajectories, metric depth, full 6DoF poses, and editable high-fidelity
        3D scene assets.
\end{itemize}

\section{Related Work}
\subsection{Panoramic Video Generation and Scene Exploration}
Panoramic content generation has progressed from static panoramic imagery toward
dynamic and explorable scenes. Early holistic multi-view image synthesis such as
PanoFree~\cite{liuPanoFreeTuningFreeHolistic2025} produces globally consistent
panoramas without per-scene tuning. To bring motion to panoramas, a prominent line of
work lifts pretrained perspective video diffusion models into the 360$^\circ$ domain:
PanoWan~\cite{NEURIPS2025_d9a12369} introduces latitude/longitude-aware mechanisms to
cope with equirectangular distortion and boundary continuity,
ViewPoint~\cite{NEURIPS2025_124ec4ed} repurposes pretrained diffusion priors for
panoramic video, and PanFlow~\cite{zhangPanFlowDecoupledMotion2026} decouples motion
control for panoramic generation. DynamicScaler~\cite{Liu_2025_CVPR} targets seamless,
scalable panoramic scene video, while 4K4DGen~\cite{ICLR2025_fa41e9d5} pushes panoramic
4D generation to 4K resolution and CubeComposer~\cite{Li_2026_CVPR} autoregressively
composes 4K 360$^\circ$ video from perspective input.
These methods produce panoramic imagery or video as the final output, with motion
driven by free or text-guided sampling rather than a prescribed camera path.

More recently, a cluster of
panoramic world models pursues large-scale, consistent scene synthesis: concurrent
PanoWorld variants address whole-house panorama
synthesis~\cite{jiaPanoWorldGenerativeSpatial2026}, geometry-consistent panoramic
video~\cite{jiangPanoWorldGeometryConsistentPanoramic2026}, and spatial
supersensing~\cite{wangPanoWorldSpatialSupersensing2026}, and
PanoWorld-X~\cite{yinPanoWorldXGeneratingExplorable2025} generates explorable panoramic
worlds via sphere-aware video diffusion.
A parallel line of work conditions video diffusion on explicit camera motion to enable
viewpoint control and scene exploration. For perspective video, GenEx~\cite{ICLR2025_8204f54f}
generates an explorable world from imagery, and Voyager~\cite{huangVoyagerLongRangeWorldConsistent2025}
produces long-range, world-consistent video for explorable 3D scene generation. A growing
family couples camera-guided generation with geometry: Matrix-3D~\cite{yangMatrix3DOmnidirectionalExplorable2025}
warps a reconstructed scene mesh to condition omnidirectional, explorable 3D world
generation, WorldStereo~\cite{Zhang_2026_CVPR} bridges camera-guided video generation and
scene reconstruction through 3D geometric memories, and large multi-modal world models such
as HY-World~2.0~\cite{hy-worldHYWorld20MultiModal2026} jointly reconstruct, generate, and
simulate 3D worlds. In the panoramic regime, OmniRoam~\cite{liuOmniRoamWorldWandering2026},
an image-to-video model built on the Wan2.1-T2V-1.3B backbone,
generates long-horizon panoramic video for world wandering, but encodes camera motion as a
scalar speed signal rather than metric $\mathrm{SE}(3)$ poses, forfeiting the spatial
grounding needed to plan and follow collision-free paths. Our panoramic video module instead injects a
dense, metric geometry-warped conditioning signal derived from a NavMesh-planned
$\mathrm{SE}(3)$ trajectory, so the generated sequence follows the requested collision-free
path while inpainting large disocclusions, which we find essential for the downstream
feed-forward 3D reconstruction.

A second gap is resource cost. These systems rely on large backbones with many-step,
classifier-free-guided sampling; methods such as WorldStereo~\cite{Zhang_2026_CVPR},
Lyra~2.0~\cite{shen2026lyra20explorablegenerative}, and
HY-World~2.0~\cite{hy-worldHYWorld20MultiModal2026} target high-end multi-GPU servers and
are hard to deploy on commodity hardware. We trade backbone scale for deployability:
shortcut-model training enables guidance-free sampling in a few denoising steps,
and the full single-panorama-to-scene pipeline runs feed-forward on a single consumer-grade GPU.

\subsection{Feed-Forward 3D Reconstruction and Gaussian Splatting}
Feed-forward 3D reconstruction has recently emerged as a promising paradigm for scalable scene reconstruction,
aiming to directly infer 3D geometry and appearance from single or sparse views in a single forward pass,
without per-scene optimization. Compared with optimization-based neural rendering methods such as NeRF and
its variants, feed-forward approaches significantly reduce reconstruction time while improving deployment
efficiency, making them particularly suitable for real-time and large-scale applications.

\subsubsection{Depth Foundation Models for 3D Reconstruction}

A key enabler of feed-forward reconstruction is the advancement of large-scale depth foundation models.
Recent methods, such as Depth Anything V3~\cite{depthanything3}, demonstrate strong generalization ability across diverse scenes
by leveraging large-scale heterogeneous training data and self-supervised objectives. These models produce
high-quality, dense depth maps that serve as robust geometric priors for downstream 3D reconstruction tasks.
In modern pipelines, such depth priors are often integrated either as explicit supervision or as initialization
for lifting 2D observations into 3D representations.

\subsubsection{Feed-Forward Multi-View Reconstruction}

Beyond monocular depth estimation, recent works focus on directly learning multi-view geometry in a
feed-forward manner. Transformer-based architectures such as VGGT~\cite{wang2025vggt} and its improved variant VGGT-$\Omega$~\cite{wang2026vggtomega}
jointly model camera pose estimation, multi-view correspondence, and dense reconstruction within a
unified network. By leveraging global attention mechanisms, these methods are capable of reasoning
about cross-view consistency without relying on traditional structure-from-motion pipelines.

Compared with earlier sparse-view reconstruction methods, VGGT-style models significantly improve
robustness under limited viewpoints and ambiguous geometry. However, they still face challenges
in high-frequency detail preservation and photorealistic rendering quality.

\subsubsection{Gaussian Splatting and Feed-Forward Radiance Representation}

Recently, 3D Gaussian Splatting (3DGS) has emerged as an efficient alternative to implicit neural
radiance fields, enabling high-quality real-time rendering through explicit Gaussian primitives.
This representation has inspired a new line of feed-forward reconstruction methods that directly
predict or initialize Gaussian parameters from images.

To improve geometric fidelity and rendering sharpness, methods such as SHARP~\cite{Sharp2025:arxiv} introduce learning-based
refinement mechanisms for Gaussian primitives, reducing over-smoothing artifacts and enhancing structural
consistency. These approaches demonstrate that Gaussian representations are particularly
well-suited for feed-forward pipelines due to their explicit nature and differentiability.

\subsubsection{Generative Priors and Multi-View Consistency}

In parallel, recent generative foundation models have begun to influence 3D reconstruction.
For example, Hunyuan-Mirror incorporates multi-view generative priors to enforce geometric
and photometric consistency across views, improving structural coherence in reconstructed scenes.
Such approaches highlight a growing trend of integrating diffusion-based or generative priors with explicit 3D
representations, bridging the gap between 2D generative modeling and 3D scene understanding.

\subsubsection{Discussion}

Overall, feed-forward 3D reconstruction is rapidly evolving toward a unified paradigm that combines:
\begin{itemize}
  \item strong depth foundation models (e.g., Depth Anything V3),
  \item multi-view transformer reasoning (e.g., VGGT / VGGT-$\Omega$),
  \item and explicit Gaussian representations (e.g., 3DGS and SHARP~\cite{Sharp2025:arxiv}),
\end{itemize}
while increasingly incorporating generative priors (e.g., Hunyuan-Mirror) to improve global consistency.

Despite these advances, current methods still struggle with fine-grained geometry recovery,
view-consistent texture synthesis, and robustness under extreme viewpoint sparsity, leaving
significant room for future improvements.

\section{Method}
\subsection{Problem Formulation}
We address the task of recovering a freely roamable 3D scene from a single
panoramic observation. The input is one equirectangular (ERP) panorama
$I\in\mathbb{R}^{H\times W\times 3}$, and the desired output is a 3D Gaussian scene model
\begin{equation}
  \mathcal{G}=\{(\mu_k,\,q_k,\,s_k,\,\alpha_k,\,c_k)\}_{k=1}^{K},
\end{equation}
where each primitive carries a position $\mu_k\in\mathbb{R}^3$, orientation $q_k$, scale
$s_k$, opacity $\alpha_k$, and view-dependent color $c_k$, enabling rasterization-based
real-time free-viewpoint rendering. A single panorama observes the scene from one vantage
point and leaves the geometry and appearance behind occluders unconstrained, so directly
regressing $\mathcal{G}$ from $I$ is ill-posed.

We therefore factorize the mapping $I\!\rightarrow\!\mathcal{G}$ into two feed-forward
stages bridged by an explicit multi-view representation. First, a
trajectory-controllable panoramic video generator
$\mathcal{F}_{\text{gen}}$ synthesizes a temporally coherent panoramic video that walks the
scene along an automatically planned, collision-free roaming trajectory
$\mathcal{T}=\{T_i\}_{i=1}^{N}$ with $T_i\in \mathrm{SE}(3)$:
\begin{equation}
  \mathcal{V}=\{V_i\}_{i=1}^{N}=\mathcal{F}_{\text{gen}}(I,\,\mathcal{T}),
\end{equation}
turning the single-view input into a set of consistent virtual observations that reveal the
initially occluded content. Second, a feed-forward panoramic reconstructor
$\mathcal{F}_{\text{rec}}$ consumes the generated video to estimate per-frame camera poses
$\{\hat{T}_i\}_{i=1}^{N}$ and a dense geometry, which a Gaussian-splatting decoder
$\mathcal{F}_{\text{dec}}$ then converts into the final scene model:
\begin{equation}
  \mathcal{G}=\mathcal{F}_{\text{dec}}\big(\mathcal{F}_{\text{rec}}(\mathcal{V})\big).
\end{equation}
Both stages are feed-forward and require no per-scene optimization: the generation
stage supplies the multi-view evidence that makes reconstruction well-posed, and the
trajectory $\mathcal{T}$ links the two, since it must be metrically faithful enough for the
reconstructor to recover consistent poses and coverage. The remainder of this section
details the trajectory-controllable panoramic video generator
(Sec.~\ref{sec:panovideo}) and the feed-forward 3D world reconstructor that produces
$\mathcal{G}$.

\begin{figure*}[t]
  \centering
  \includegraphics[width=\textwidth]{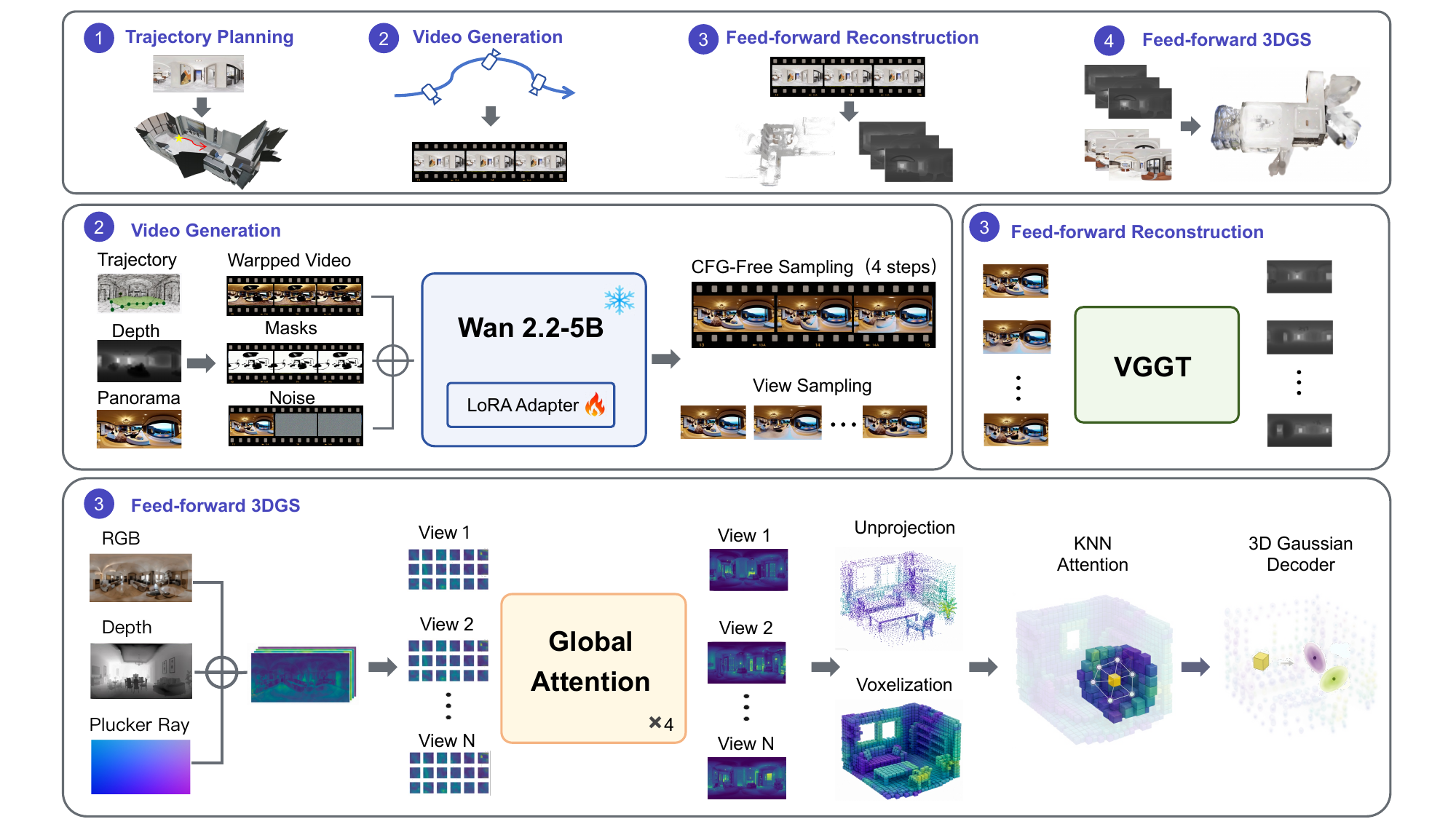}
  \caption{\textbf{Overview of Genie Sim PanoWorld.}
  \emph{Stage~1 (Environment Perception \& Trajectory Construction):} A raw ERP panorama is
  processed by a depth estimator to build a dense point cloud and 3D mesh; a NavMesh is
  extracted from the traversable floor region and used to plan a collision-free
  $\mathrm{SE}(3)$ camera trajectory.
  \emph{Stage~2 (Latent Video Diffusion):} For each planned pose, a geometry-warped ERP
  frame and its disocclusion mask are encoded by a frozen 3D VAE and concatenated with the
  noisy latent along the channel axis.  A Latent Rectified-Flow DiT (Wan2.2-TI2V-5B + LoRA)
  denoises the sequence in 4 CFG-free steps via shortcut-model sampling; the
  frozen VAE decoder produces the final $T$-frame panoramic video.
  \emph{Stage~3 (Feed-Forward 3D Reconstruction \& Voxel-Based Gaussian Splatting):} The
  generated panoramic frames are fed to PanoVGGT, a geometry transformer that jointly
  estimates camera poses, dense depth maps, and high-resolution image features.
  Observations are lifted into world coordinates, aggregated into occupied voxels via
  multi-view attention and KNN attention blocks, and decoded per voxel into 3D Gaussian
  primitives for real-time free-viewpoint rendering.}
  \label{fig:overview}
\end{figure*}

\subsection{Trajectory-Controllable Panoramic Video Generation}
\label{sec:panovideo}

Given a single ERP panorama $I\in\mathbb{R}^{H\times W\times 3}$, the goal of this stage
is to synthesize a temporally coherent panoramic video
$\mathcal{V}=\{V_i\}_{i=1}^{N}$ that follows an automatically planned, collision-free
roaming trajectory $\mathcal{T}=\{T_i\}_{i=1}^{N}$, $T_i\in \mathrm{SE}(3)$, and that remains
photometrically consistent with $I$. The trajectory is planned on a NavMesh derived from
the input panorama's depth, which keeps every camera pose within the navigable floor
region and prevents wall penetration. In contrast, OmniRoam~\cite{liuOmniRoamWorldWandering2026}
conditions on a scalar speed signal and therefore does not preserve the metric spatial layout
required for collision-aware path control. Our generator is a trajectory-conditioned latent
video diffusion model built on the Wan2.2-TI2V-5B backbone~\cite{wan2025wan22}, adapted
with low-rank adapters~\cite{hu2022lora}. Matrix-3D~\cite{yangMatrix3DOmnidirectionalExplorable2025}
relies only on geometric warps and degrades under long indoor camera displacements. Our
generator keeps the geometry-warped conditioning but adds two changes: (i) it trains on a
mixture of long- and short-range trajectories that couples high-fidelity supervision
with large-disocclusion inpainting, and (ii) it applies a self-consistency objective based on
shortcut models to enable high-quality, few-step sampling without classifier-free guidance
(CFG), so the model can run on a single consumer GPU.

\subsubsection{NavMesh-Guided Trajectory Planning and Conditioning}
\label{sec:warp_cond}
We first recover the coarse 3D structure of the input scene. A panoramic depth map
$D\in\mathbb{R}^{H\times W}$ is estimated with Depth Anything
360$^\circ$~\cite{jiang2025depth360circscaleinvariance}, back-projected into a dense point
cloud, and triangulated into a scene mesh $\mathcal{M}$. We extract the traversable floor
region from $\mathcal{M}$ and build a NavMesh as the free-space prior for collision-free
path planning.

On the NavMesh we automatically generate a roaming trajectory
$\mathcal{T}=\{T_i\}_{i=1}^{N}$ ($N=81$ frames) by combining \emph{sector exploration
paths}, which cover the full 360° extent of the scene, and \emph{ring paths}, which form
closed-loop tours that emphasise temporal consistency. All paths are re-parameterized to a
near-uniform step length and expressed in a canonical frame anchored at the first view.

We then convert the planned trajectory into a dense conditioning signal: for each pose
$T_i$ we render the scene mesh into ERP space to obtain a warped panorama $\hat{V}_i$ and
a disocclusion mask $M_i$, which are jointly encoded and injected into the DiT, turning
the generator into a geometry-aware video inpainter.

\subsubsection{Mixed Training and Self-Consistency Loss}
\label{sec:longshort}
\label{sec:shortcut}
\paragraph{Long--short trajectory mixed training.}
The geometry-warped condition degrades with camera displacement: long traversals produce
large disocclusions that geometry-only conditioning cannot fill, while short-range warps
are near-perfect but provide no inpainting supervision. We therefore train on a
mixture of long- and short-range clips drawn from the same ground-truth tracks,
with clip lengths sampled uniformly so that every traversal scale is equally represented.
Short-range clips anchor appearance fidelity; long-range clips supervise large-disocclusion
inpainting.

\paragraph{Shortcut-model training for CFG-free sampling.}
The model is a flow-matching (rectified-flow) diffusion in latent
space~\cite{lipman2023flowmatching}. With noise level $\sigma=t/N_{\!t}$, the noisy
latent is $x_t=(1-\sigma)\,x_0+\sigma\,\epsilon$ and the velocity target is
$v=\epsilon-x_0$.

To reduce the number of inference steps, we adopt the self-consistency formulation of
shortcut models~\cite{frans2025oneshortcut}, which additionally conditions the network on a step
size $\Delta t$: $s_\theta(x_t,t,\Delta t,y,c)$. Rather than retraining the full 5B
backbone, we apply LoRA fine-tuning with the self-consistency objective, which removes the
need for a CFG unconditional branch entirely. Each iteration is routed with probability $\rho=0.75$ to the
flow-matching loss
\begin{equation}
  \mathcal{L}_{\mathrm{FM}}=\big\|\,s_\theta(x_t,t,0)-v\,\big\|^2,
\end{equation}
and with probability $1-\rho$ to the self-consistency loss, where a frozen teacher takes
two half-steps to produce a more accurate velocity estimate $\bar{v}=\tfrac{1}{2}(v_1+v_2)$
and the student must match it:
\begin{equation}
  \mathcal{L}_{\mathrm{SC}}=\big\|\,s_\theta(x_t,t,\Delta t)-\bar{v}\,\big\|^2.
\end{equation}
At inference, $4$ large Euler steps with $w=1$ (no CFG) suffice, substantially reducing
NFE relative to the Matrix-3D baseline. We evaluate Matrix-3D using $10$ guided steps,
a practical setting that balances generation quality and inference efficiency, instead
of its full $50$-step schedule. A lightweight RealESRGAN~\cite{wang2021realesrgan} $\times2$ stage finally
upscales the decoded frames.

\paragraph{Training objective.}
The video generator is trained independently from the reconstruction stage:
\begin{equation}
  \mathcal{L}_{\text{video}}
   = b\,\mathcal{L}_{\mathrm{FM}}
   + (1-b)\,\mathcal{L}_{\mathrm{SC}},
\end{equation}
where $b\sim\mathrm{Bernoulli}(\rho)$. Only the LoRA adapters, the conditioning
projection, and the step-size embedding are updated; the Wan2.2 backbone and VAE
remain frozen.

\subsection{Feed-Forward 3D World Reconstruction}
\label{sec:recon}

\subsubsection{Feed-Forward Pose and Depth Estimation}

We build upon recent feed-forward multi-view reconstruction frameworks, particularly VGGT,
and extend them to panoramic 3D world reconstruction. Unlike conventional perspective-based
formulations, which rely on pinhole camera assumptions and explicit intrinsic parameter
estimation, we reformulate the problem in an equirectangular projection (ERP) domain to
enable holistic scene understanding from a single panoramic observation.

A key motivation for adopting panoramic ERP representation is its superior spatial efficiency
in indoor and confined environments. In narrow indoor scenes, a single panoramic observation
can capture the entire surrounding geometry in a unified coordinate system, significantly
reducing viewpoint fragmentation and improving global geometric completeness. Compared with
multi-view perspective setups, ERP inherently encodes 360° contextual information, which
provides stronger constraints for pose inference and reduces ambiguity caused by limited
field-of-view observations.

Specifically, instead of processing multiple perspective views with separate camera intrinsics,
we directly operate on ERP images. This design eliminates the need for explicit camera
intrinsic estimation, since ERP projection provides a fixed and consistent mapping from
spherical directions to the image plane. As a result, the model can focus on learning global
scene geometry and pose relations without disentangling per-view calibration parameters,
thereby simplifying the learning objective and improving optimization stability.

This reformulation also benefits pose estimation accuracy. In conventional perspective-based
pipelines, pose estimation is often ill-posed in textureless or geometrically ambiguous indoor
regions due to limited visibility. In contrast, ERP-based observation provides globally
consistent angular coverage, enabling stronger cross-region geometric constraints and improving
robustness of relative pose prediction. Empirically, this leads to more stable and accurate
pose estimation, particularly in indoor scenes with repetitive structures or narrow corridors.

A further key modification over VGGT lies in the positional encoding strategy. Existing
VGGT-style models rely on perspective-view 3D positional embeddings to encode spatial structure
across views. However, such representations are not well-suited for panoramic geometry due to
discontinuities and distortions introduced by perspective projection. In contrast, we remove
spherical 3D positional encoding and instead adopt a direct ERP-aligned coordinate encoding,
which provides a continuous and globally consistent representation of angular coordinates on
the panoramic image plane.

This reformulation enables the model to implicitly reason about full-scene geometry in a
unified spherical observation space, improving cross-region consistency and reducing ambiguity
at image boundaries. Moreover, by eliminating the need for intrinsic parameter regression and
spherical embedding design, the proposed framework simplifies the overall reconstruction
pipeline while maintaining strong geometric reasoning capability.

\subsubsection{Feed-Forward 3D Gaussian Splatting}
\label{sec:ff_3dgs}

Given the estimated panoramic poses and dense point maps from the feed-forward reconstructor, we decode a renderable 3D Gaussian scene without any per-scene optimization. A straightforward strategy would be to regress one Gaussian from each image pixel independently. However, such pixel-aligned predictions are highly redundant across overlapping views and tend to produce view-dependent primitives. We instead adopt a voxel-aligned Gaussian head that lifts multi-view panoramic observations into a shared 3D volume, fuses them into occupied voxels, and predicts Gaussian primitives from the resulting 3D tokens.

For each context panorama $V_i$, the feed-forward network estimates a dense depth map and a camera pose $\hat{T}_i$. We compute the Pl\"ucker ray encoding for every pixel $u$, which compactly represents both the ray origin $o_i$ and the normalized ray direction $d_{i,u}$. The input feature map is then formed by concatenating the RGB color, the predicted depth, and the Pl\"ucker ray encoding along the channel dimension:
\begin{equation}
  f_{i,u} = [I_{i,u},\; z_{i,u},\; d_{i,u},\; o_i \times d_{i,u}],
\end{equation}
where $I_{i,u}$ denotes the pixel color and $z_{i,u}$ is the predicted depth at pixel $u$.

To capture global structure and cross-view correlations, the feature maps from all context views are partitioned into non-overlapping patches and fed into a global attention module. Self-attention is performed jointly over all patches from all views, enabling the network to reason about inter-view correspondences and resolve geometric ambiguities. After attention, the processed patches are reassembled into per-view feature maps. To minimize the spatial gap between neighboring pixels and preserve fine-grained detail, we upsample the feature maps back to the original panoramic resolution before back-projection.

Each pixel feature is then back-projected into 3D world space using the estimated depth and pose. We quantize the resulting 3D points into a voxel grid with cell size $\epsilon$ and aggregate all observations falling into the same voxel. To account for varying observation quality, the network predicts a per-pixel confidence weight $w_{i,u}$, yielding a weighted voxel feature:
\begin{equation}
  f^{\mathrm{vox}}_j = \frac{\sum_{(i,u)\in\mathcal{S}_j} w_{i,u}\, f_{i,u}}{\sum_{(i,u)\in\mathcal{S}_j} w_{i,u} + \varepsilon}, 
\end{equation}
\begin{equation}
    p_j = \frac{\sum_{(i,u)\in\mathcal{S}_j} w_{i,u}\, x_{i,u}}{\sum_{(i,u)\in\mathcal{S}_j} w_{i,u} + \varepsilon},
\end{equation}
where $\mathcal{S}_j$ denotes the set of observations assigned to voxel $j$, $x_{i,u}$ is the 3D position of pixel $u$ in world coordinates, and $p_j$ is the resulting voxel center. This fusion step converts redundant multi-view pixel observations into compact, view-consistent 3D tokens.

To inject 3D structural priors into the fused representations, we apply a $k$-nearest-neighbor (KNN) attention module over the occupied voxel tokens. Each voxel attends to its spatial neighbors in 3D, allowing the network to capture local geometric relationships and propagate contextual information across adjacent regions, thereby enriching the 2D-originated features with explicit 3D spatial structure.

Each occupied voxel token, augmented with a positional encoding of its center $p_j$, is decoded by a compact MLP into $K$ Gaussian primitives. For the $k$-th primitive of voxel $j$, the network predicts a center offset, scale, rotation quaternion, opacity, and spherical-harmonic color coefficients:
\begin{equation}
  (\Delta\mu_{j,k},\; s_{j,k},\; q_{j,k},\; \alpha_{j,k},\; c_{j,k}) = \phi_{\mathrm{gs}}\big([f^{\mathrm{vox}}_j,\; \gamma(p_j)]\big),
\end{equation}
where $\gamma(\cdot)$ denotes sinusoidal positional encoding. The predicted attributes are parameterized with bounded activations for stable feed-forward inference. The final Gaussian center is obtained as $\mu_{j,k} = p_j + \Delta\mu_{j,k}$, and the complete scene is represented as
\begin{equation}
  \mathcal{G} = \{(\mu_k, q_k, s_k, \alpha_k, c_k)\}_{k=1}^{N_g},
\end{equation}
which can be rendered in real time via differentiable Gaussian rasterization.


\subsubsection{Training}

\paragraph{Training Losses}
The feed-forward reconstruction model is trained in two stages.
 In the first stage, we optimize the camera pose head and depth
  head using only the camera pose loss and depth loss, following 
  the original formulation of VGGT~\cite{wang2025vggt} without 
  modification. Specifically, the camera loss $\mathcal{L}_{\text{camera}}$ supervises 
  the predicted relative camera poses between frame pairs, while the depth loss
   $\mathcal{L}_{\text{depth}}$ applies scale-invariant log-depth regression to the
    predicted dense point maps. The first-stage objective is
\begin{equation}
\mathcal{L}_{\text{stage1}}
=
\mathcal{L}_{\text{camera}}
+
\mathcal{L}_{\text{depth}}.
\end{equation}

In the second stage, we freeze the trained camera pose and depth heads and 
optimize the feed-forward 3D Gaussian Splatting task head. The predicted 
3D Gaussians are rendered into the input views, and the task head is supervised 
using an RGB reconstruction loss implemented as an $\ell_2$ loss between the 
rendered and ground-truth images:
\begin{equation}
\mathcal{L}_{\text{stage2}}
=
\mathcal{L}_{\text{gs}}
=
\left|
\hat{\mathbf{I}}-\mathbf{I}
\right|_2^2,
\end{equation}
where $\hat{\mathbf{I}}$ and $\mathbf{I}$ denote the rendered and ground-truth 
RGB images, respectively. This staged training strategy first establishes reliable 
geometric predictions and subsequently learns to convert them into feed-forward 3D 
Gaussian scene representations.

\paragraph{Training Data}
Our model is trained on three panoramic indoor datasets. \textbf{RealSee3D}~\cite{Li2025realsee3d_data} 
provides $9{,}000$ procedurally-generated synthetic indoor scenes with a total 
of $270$K viewpoints, featuring diverse layouts and decoration styles.
 \textbf{InteriorGS}~\cite{miao2025physicallyexecutable3dgaussian} provides $602$ real-world
 indoor scenes, from which we uniformly sample $50$ frames per scene from $800$-frame dense 
 trajectories for model training. \textbf{Structured3D}~\cite{zheng2020structured3d} offers $3{,}131$ synthetic 
 indoor scenes with photorealistic rendering and rich structural annotations. All datasets provide ground-truth 
 camera poses and depth maps, which serve as supervision for the feed-forward reconstruction model.

\newcommand{\OURS}{\textbf{PanoHome}}   
\newcommand{\yes}{\checkmark}
\newcommand{\no}{\texttimes}
\begin{table*}[t]
\centering
\setlength{\tabcolsep}{5pt}
\scriptsize
\caption{Comparison with prior \emph{indoor} panoramic datasets
(\checkmark: available; \texttimes: unavailable; $\sim$: partial / restricted).
Unlike existing datasets that are sparse, geometrically imprecise, or provide no
editable 3D, our procedurally generated dataset offers pixel-perfect metric
geometry, dense per-room multi-view overlap, and a complete editable 3D scene
for every sample.}
\vspace{-2mm}
\label{tab:datasets}
\begin{tabular}{l c c c c c l}
\toprule
Dataset & \#Images & Real/Syn. & Pose & Depth & 3D Model & Characteristics \\
\midrule
PanoSUNCG~\cite{wang2018self}      & $\sim$25K  & Syn.  & 6DoF & 8-bit              & \no          & Deprecated, not legally available \\
Matterport3D~\cite{chang2017matterport3d} & $\sim$10.8K & Real & 6DoF & 16-bit$^{\dagger}$ & $\sim$ recon. & Discrete panos, limited overlap \\
Structured3D~\cite{zheng2020structured3d} & $\sim$12K & Syn. & 3DoF & 16-bit            & $\sim$ CAD    & One pano per room, no overlap \\
Stanford2D3D~\cite{armeni2017joint}& $\sim$1.4K & Real  & 6DoF & 16-bit$^{\dagger}$ & \yes mesh    & Small scale, limited overlap \\
Pano3D~\cite{albanis2021pano3d}    & $\sim$42K  & Mixed & \no  & 16-bit             & \no          & No camera pose \\
360Loc~\cite{huang2024360loc}      & $\sim$2.9K & Real  & 6DoF & sparse$^{\dagger}$ & \no          & LiDAR depth, RGB--depth mismatch \\
\midrule
\OURS                              & $\sim$30K  & Syn.  & 6DoF & float32            & \yes \emph{editable} & Dense ($\sim$30 panos/room), high overlap \\
\bottomrule
\end{tabular}
\vspace{-1mm}
\\[2pt]
{\scriptsize $^{\dagger}$ Real captured depth: contains sensor noise, missing regions, or RGB--depth misalignment.}
\vspace{-3mm}
\end{table*}

\vspace{-0.4em}
To mitigate the scarcity of high-quality synthetic data, we generate our own large-scale dataset,
\OURS, a photorealistic indoor panoramic dataset. We build on the
procedural scene generators Infinigen~\cite{infinigen2023infinite} and Infinigen
Indoors~\cite{infinigen2024indoors}, which synthesize fully parametric,
physically based 3D rooms; our contribution replaces their perspective camera
with an equirectangular ($360^\circ$) panoramic renderer and a per-room dense
trajectory sampler tailored to feed-forward panoramic reconstruction. Because
every scene is synthesized rather than captured, RGB, depth, and pose are
rendered from the same ground-truth geometry and are therefore \emph{perfectly
aligned}, free of the sensor noise, holes, and privacy or licensing constraints
of real captures. Each sample provides an equirectangular RGB image
($2880\times1440$) and a \emph{continuous
metric depth map} at \texttt{float32} precision—strictly finer than the $16$-bit
quantized depth of prior datasets—together with a complete $6$DoF
camera-to-world pose (the equirectangular parameterization fully determines the
projection, so no pinhole intrinsics are needed). We further sample $\sim\!30$
panoramas per room along controlled trajectories, yielding \emph{dense
multi-view overlap}, and release every scene as a complete, \emph{editable} 3D
asset that enables re-rendering, additional supervision, and embodied simulation.
As summarized in~\Cref{tab:datasets}, these properties directly address the
limitations of existing indoor panoramic datasets.
Matterport3D~\cite{chang2017matterport3d} and
Stanford2D3D~\cite{armeni2017joint} provide only discrete panoramas with limited
overlap; Structured3D~\cite{zheng2020structured3d} renders a single panorama per
room, yielding \emph{no} cross-view overlap; Pano3D~\cite{albanis2021pano3d}
omits camera poses; and 360Loc~\cite{huang2024360loc} suffers from RGB--depth
misalignment. Crucially, none of them exposes an editable 3D scene, whereas
\OURS{} combines dense, high-overlap trajectories with pixel-perfect metric
geometry and full 3D editability, and its procedural nature makes scale and
diversity effectively unbounded.

\section{Experiments}
\subsection{Panoramic Video Generation Data}
Our experiments use two panoramic sources.
\textbf{InteriorGS}~\cite{InteriorGS2025} provides $602$ photorealistic indoor scenes
with RGB panoramas, dense ERP camera trajectories of up to $800$ frames per scene, and
ground-truth depth from SAGE-3D~\cite{miao2025physicallyexecutable3dgaussian}. For each
scene we build trajectory-conditioned training tuples: the ground-truth camera track
defines the roaming trajectory $\mathcal{T}$, and we render the geometry-warped condition
video and its visibility mask from the scene's panorama and depth. Clip lengths are sampled
uniformly so that a single scene yields many clips spanning a continuum of traversal
scales. All conditions are pre-rendered offline and cached so that training is I/O-bound
rather than render-bound. The held-out test split contains $60$ scenes for ground-truth
evaluation.
\textbf{RealSee3D}~\cite{Li2025realsee3d_data} provides $9{,}000$ procedurally-generated
synthetic indoor scenes with diverse layouts and decoration styles. Since it provides only
static RGB-D panoramic captures without roaming video sequences, we use it solely for
zero-shot evaluation, drawing $100$ panoramic scenes as the generalization set.

\subsection{Metrics}
\textbf{Panoramic video generation.} On the InteriorGS test split we report frame-level
fidelity (PSNR, SSIM, LPIPS, WS-PSNR) and video-level realism (FVD, FID); FVD is used
for relative comparison across models only, as the test set is too small for an
absolute score. To quantify
trajectory controllability, we reconstruct camera poses from the generated frames via
COLMAP, align to the ground-truth trajectory with $\mathrm{Sim}(3)$ Umeyama, and report
ATE, RRE, and RTE. On RealSee3D, which has no roaming-video ground truth, we drive
inference with our own collision-free ring trajectories and report FID, a loop-closure
score (following OmniRoam~\cite{liuOmniRoamWorldWandering2026}), first/last-frame
LPIPS-Loop, and the same pose metrics against the planned trajectory.

\textbf{3D reconstruction.}
We evaluate the reconstruction module along two axes.
\emph{(i) Trajectory estimation accuracy.}
Predicted camera poses are $\mathrm{Sim}(3)$-aligned to the ground-truth trajectory via the
Umeyama algorithm, and we report the Absolute Trajectory Error (ATE), its RMSE variant
(ATE-RMSE), and the Relative Pose Error in translation (RPE$_t$) and rotation (RPE$_r$,
in degrees) computed between consecutive frame pairs.
\emph{(ii) Rendering quality.}
We render the reconstructed 3D Gaussians from both the input (context) camera poses and
held-out novel viewpoints, and report PSNR, SSIM, and LPIPS against the corresponding
ground-truth panoramas.

\subsection{Implementation Details}
\textbf{Panoramic video generation.}
We adapt Wan2.2-TI2V-5B~\cite{wan2025wan22} with LoRA~\cite{hu2022lora}
(rank $48$, $\alpha=48$) initialized from Matrix-3D~\cite{yangMatrix3DOmnidirectionalExplorable2025}.
The backbone and 3D VAE are frozen; only the LoRA adapters, geometry-warp conditioning tokens,
and the step-size embedding are trained. All models are optimized with AdamW in bf16
using DeepSpeed ZeRO-1 and gradient checkpointing, at $720\!\times\!1440$ with $81$ frames.
For ablation we train a +Mixed Training variant using only the flow-matching objective
without the shortcut-model formulation. At inference, the shortcut model uses $4$ CFG-free
denoising steps; the ablation baseline uses $10$ guided steps.\\
\textbf{Feed-forawrd reconstruction.}
We resize equirectangular panoramas of varying resolutions to $392 \times 784$ for training. The model is trained 
on $32$ NVIDIA H100 (80\,GB) GPUs for $8$ days. For each batch, 
we randomly sample $2$ to $32$ frames from a randomly selected training scene, 
enabling the model to handle varying numbers of context views at inference time.
 All training is conducted in bf16 mixed precision with the AdamW optimizer, 
 following the hyperparameter settings of the original VGGT~\cite{wang2025vggt}.


\subsection{Panoramic Video Generation}

\begin{table*}[t]
\centering
\caption{Panoramic video generation on InteriorGS ($60$ scenes), merging frame-level fidelity
and trajectory controllability under both protocols. \emph{Top block:} native
$720\times1440$. \emph{Bottom block:} outputs from all methods are resized to a common resolution of
$480\times960$ before computing the evaluation metrics. This block also includes
OmniRoam~\cite{liuOmniRoamWorldWandering2026} ($10$ steps, CFG scale $5.0$). Pose metrics (ATE, RRE, RTE) are estimated via
COLMAP and $\mathrm{Sim}(3)$-aligned to the ground-truth trajectory. \emph{GT track} = poses
re-estimated from the ground-truth frames via COLMAP (reference).
$\uparrow$/$\downarrow$ = higher/lower is better; \textbf{bold} = best per trajectory stride within each
block (excluding reference).}
\label{tab:video_main}
\footnotesize
\setlength{\tabcolsep}{3.5pt}
\begin{tabular}{llccccccccccc}
\toprule
 & Method & Stride $r$ & Steps & PSNR$\uparrow$ & SSIM$\uparrow$ & LPIPS$\downarrow$ & WS-PSNR$\uparrow$ & FVD$\downarrow$ & FID$\downarrow$ & ATE$\downarrow$ & RRE\,($^\circ$)$\downarrow$ & RTE$\downarrow$ \\
\midrule
\multirow{9}{*}{\rotatebox[origin=c]{90}{$720\times1440$}}
 & GT track (ref.) & $r{=}1$ & -- & \multicolumn{6}{c}{--} & 0.0024 & 0.0269 & 0.0008 \\
 & Matrix-3D       & $r{=}1$ & 10  & 21.584 & 0.7977 & 0.2771 & 21.146 & 11.101 & 26.052 & 0.0081 & 0.0274 & 0.0021 \\
 & Ours (4-step)   & $r{=}1$ & \;4 & \textbf{23.499} & \textbf{0.8272} & \textbf{0.2330} & \textbf{22.970} & \textbf{7.932} & \textbf{18.249} & \textbf{0.0035} & \textbf{0.0212} & \textbf{0.0016} \\
\cmidrule(l){2-13}
 & GT track (ref.) & $r{=}2$ & -- & \multicolumn{6}{c}{--} & 0.0013 & 0.0158 & 0.0007 \\
 & Matrix-3D       & $r{=}2$ & 10  & 19.958 & 0.7562 & 0.3190 & 19.489 & 12.270 & 27.921 & 0.0312 & 0.3212 & 0.0073 \\
 & Ours (4-step)   & $r{=}2$ & \;4 & \textbf{21.925} & \textbf{0.7905} & \textbf{0.2718} & \textbf{21.367} & \textbf{9.228} & \textbf{20.310} & \textbf{0.0068} & \textbf{0.0285} & \textbf{0.0026} \\
\cmidrule(l){2-13}
 & GT track (ref.) & $r{=}4$ & -- & \multicolumn{6}{c}{--} & 0.0026 & 0.0216 & 0.0010 \\
 & Matrix-3D       & $r{=}4$ & 10  & 18.034 & 0.7064 & 0.3749 & 17.607 & 15.066 & 30.720 & 0.0952 & 0.1333 & 0.0147 \\
 & Ours (4-step)   & $r{=}4$ & \;4 & \textbf{20.288} & \textbf{0.7506} & \textbf{0.3194} & \textbf{19.746} & \textbf{10.810} & \textbf{22.133} & \textbf{0.0218} & \textbf{0.0460} & \textbf{0.0066} \\
\midrule
\multirow{12}{*}{\rotatebox[origin=c]{90}{$480\times960$}}
 & GT track (ref.) & $r{=}1$ & -- & \multicolumn{6}{c}{--} & 0.0026 & 0.0316 & 0.0014 \\
 & OmniRoam        & $r{=}1$ & 10  & 17.562 & 0.6596 & 0.3312 & 17.303 & 11.708 & 24.382 & 0.0160 & 0.0872 & 0.0034 \\
 & Matrix-3D       & $r{=}1$ & 10  & 22.003 & 0.8059 & 0.2500 & 21.615 & 10.831 & 24.319 & 0.0085 & \textbf{0.0358} & 0.0024 \\
 & Ours (4-step)   & $r{=}1$ & \;4 & \textbf{23.889} & \textbf{0.8342} & \textbf{0.2105} & \textbf{23.407} & \textbf{7.746} & \textbf{18.006} & \textbf{0.0048} & 0.0371 & \textbf{0.0023} \\
\cmidrule(l){2-13}
 & GT track (ref.) & $r{=}2$ & -- & \multicolumn{6}{c}{--} & 0.0034 & 0.0311 & 0.0013 \\
 & OmniRoam        & $r{=}2$ & 10  & 16.826 & 0.6373 & 0.3678 & 16.541 & 13.121 & 25.887 & 0.0358 & 0.1791 & 0.0115 \\
 & Matrix-3D       & $r{=}2$ & 10  & 20.306 & 0.7585 & 0.2959 & 19.880 & 11.678 & 26.676 & 0.0232 & 0.1358 & 0.0048 \\
 & Ours (4-step)   & $r{=}2$ & \;4 & \textbf{22.247} & \textbf{0.7927} & \textbf{0.2498} & \textbf{21.727} & \textbf{9.022} & \textbf{20.122} & \textbf{0.0066} & \textbf{0.0377} & \textbf{0.0028} \\
\cmidrule(l){2-13}
 & GT track (ref.) & $r{=}4$ & -- & \multicolumn{6}{c}{--} & 0.0092 & 0.1159 & 0.0034 \\
 & OmniRoam        & $r{=}4$ & 10  & 16.021 & 0.6185 & 0.4120 & 15.673 & 14.665 & 26.631 & 0.0685 & 0.2288 & 0.0187 \\
 & Matrix-3D       & $r{=}4$ & 10  & 18.198 & 0.6881 & 0.3517 & 17.795 & 15.034 & 29.885 & 0.1071 & 0.5378 & 0.0210 \\
 & Ours (4-step)   & $r{=}4$ & \;4 & \textbf{20.545} & \textbf{0.7465} & \textbf{0.2997} & \textbf{20.035} & \textbf{10.531} & \textbf{21.630} & \textbf{0.0224} & \textbf{0.1342} & \textbf{0.0070} \\
\bottomrule
\end{tabular}
\end{table*}

We evaluate trajectory-controllable panoramic video generation by comparing our full model,
Ours (4-step), against the geometry-only conditioning baseline
Matrix-3D~\cite{yangMatrix3DOmnidirectionalExplorable2025}. Both methods share the same
Wan2.2-TI2V-5B~\cite{wan2025wan22} backbone, conditions, and trajectories; they differ in the proposed training components and
the resulting sampler ($10$ guided steps for Matrix-3D vs.\ $4$ guidance-free steps for ours).
Per-component ablation is reported separately in Table~\ref{tab:video_ablation}. We evaluate
trajectory subsampling factors $r\in\{1,2,4\}$. For a fixed sequence length of $81$ frames,
$r$ denotes the temporal stride used to sample the ground-truth trajectory. Increasing $r$
therefore enlarges both the inter-frame camera displacement and the total traversal extent
while keeping the number of generated frames fixed. This evaluation targets the long-distance
regime that motivates our design.

\subsubsection{Quantitative Comparison}

The top block of Table~\ref{tab:video_main} reports frame-level fidelity and
trajectory controllability on the InteriorGS test split. Ours (4-step) outperforms Matrix-3D
at every trajectory stride by roughly $+2$\,dB PSNR with large reductions in LPIPS, FVD and FID. The gap
widens with traversal distance: at $r{=}4$ Matrix-3D drops to PSNR
$18.03$ and FID $30.72$, while our model stays higher at PSNR $20.29$ and FID $22.13$. This
indicates that the proposed training strategy mitigates the long-distance quality collapse of
geometry-only conditioning. Pose accuracy is verified independently via COLMAP reconstruction
on the generated and ground-truth frames. On trajectory accuracy, Matrix-3D's pose tracking
collapses at $r{=}4$ (ATE $0.095$ vs.\ ground-truth reference $0.003$), whereas our model
stays stable (ATE $0.022$) and is close to the ground-truth accuracy at the smaller strides.

\FloatBarrier
\subsubsection{Qualitative Comparison}

\begin{figure*}[t]
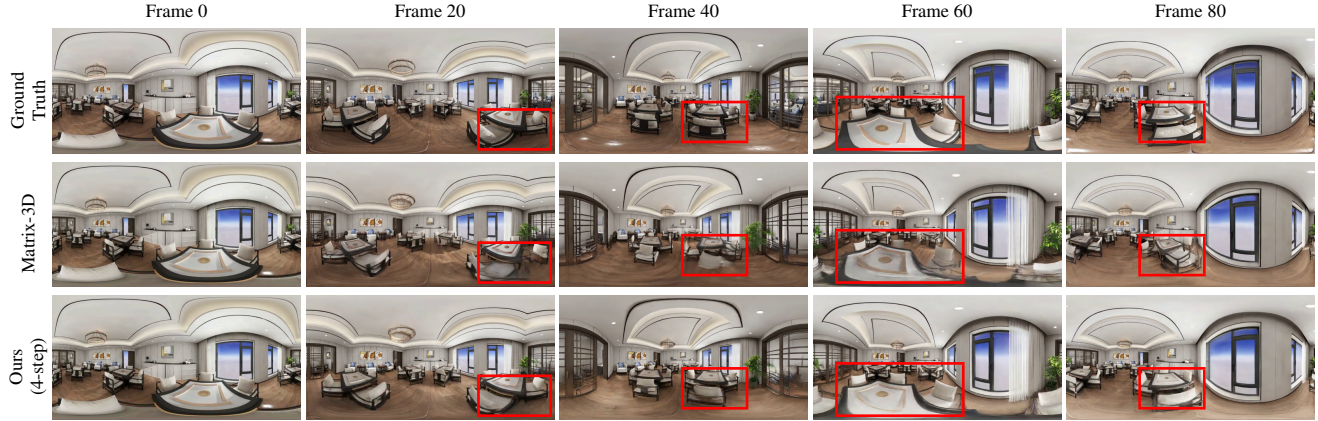

  \centering
  \setlength{\tabcolsep}{1pt}
  \renewcommand{\arraystretch}{0.95}
  \begin{tabular}{ccccccc}
    & \scriptsize Frame 0 & \scriptsize Frame 20 & \scriptsize Frame 40
    & \scriptsize Frame 60 & \scriptsize Frame 80 \\[1pt]
    \qrow{Ground\\Truth}{gt}
    \qrow{Matrix-3D}{matrix3d}
    \qrow{Ours\\(4-step)}{ours4step}
  \end{tabular}
  \caption{Qualitative comparison on a long-range ($r{=}4$) indoor traversal.
  Matrix-3D degrades into blurry, structurally inconsistent content in disoccluded regions
  as the camera travels farther, while our full model remains sharp and geometrically
  coherent. GT shown for reference; per-component ablation in Table~\ref{tab:video_ablation}.}
  \label{fig:qualitative_longrange}
\end{figure*}

Figure~\ref{fig:qualitative_longrange} compares Matrix-3D with Ours (4-step) on a
long-range ($r{=}4$) indoor traversal. As the camera moves far from the input viewpoint,
Matrix-3D produces blurred textures and broken structures in newly disoccluded regions, whereas
our model remains sharp and geometrically coherent with only $4$ denoising steps.

For a direct comparison with OmniRoam~\cite{liuOmniRoamWorldWandering2026}, whose native
output resolution differs from ours, we resize all outputs to a common $480\times960$ and recompute
every metric (bottom block of Table~\ref{tab:video_main}). OmniRoam, which is based on an
image-to-video model (Wan2.1-T2V-1.3B) and therefore uses a considerably smaller and
architecturally distinct backbone from ours, trails
both geometry-conditioned models by a large margin (PSNR $17.56$ vs.\ our $23.96$ at
$r{=}1$) and drifts severely in pose tracking as the traversal lengthens (ATE $0.069$
at $r{=}4$), similar to Matrix-3D. OmniRoam is also structurally incompatible with our
trajectory-controllable formulation: it conditions on a normalized trajectory
representation (raw poses collapsed into a scalar speed signal) that discards the metric
geometry needed to follow the collision-free NavMesh paths produced by our planner. Ours
(4-step) outperforms both baselines at every trajectory stride on all metrics.

\begin{figure*}[t]
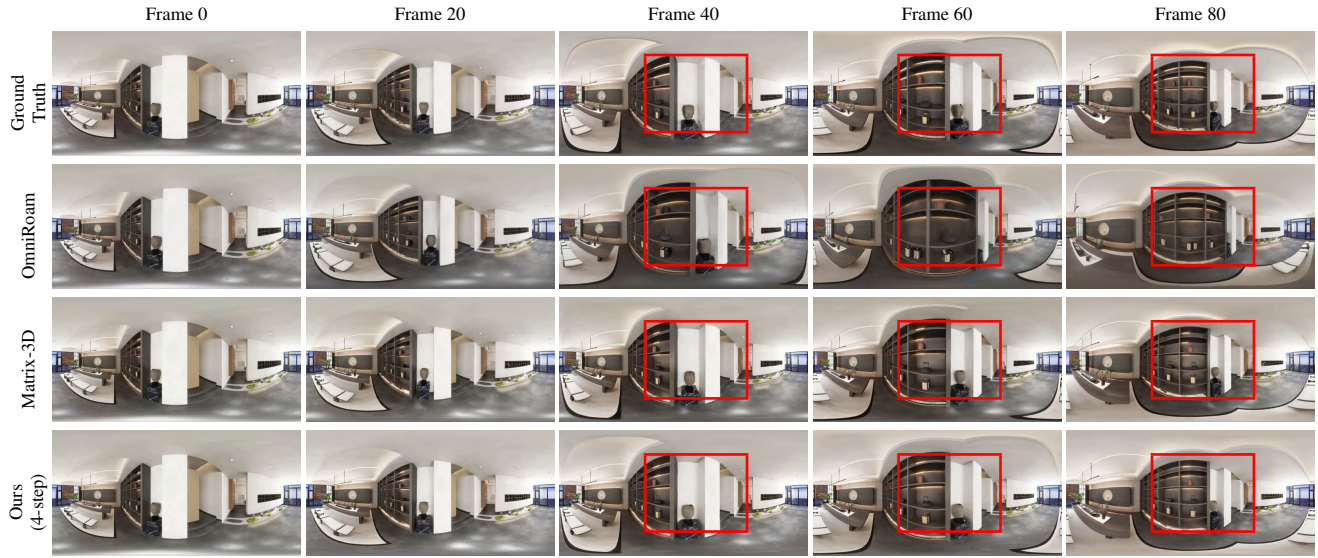

  \centering
  \setlength{\tabcolsep}{1pt}
  \renewcommand{\arraystretch}{0.95}
  \begin{tabular}{ccccccc}
    & \scriptsize Frame 0 & \scriptsize Frame 20 & \scriptsize Frame 40
    & \scriptsize Frame 60 & \scriptsize Frame 80 \\[1pt]
    \qrowGBoxLastThree{Ground\\Truth}{panovideo/omniroam_s1_0082_839951}{gt}
    \qrowGBoxLastThree{OmniRoam}{panovideo/omniroam_s1_0082_839951}{omniroam}
    \qrowGBoxLastThree{Matrix-3D}{panovideo/omniroam_s1_0082_839951}{matrix3d}
    \qrowGBoxLastThree{Ours\\(4-step)}{panovideo/omniroam_s1_0082_839951}{ours4step}
  \end{tabular}
  \caption{Qualitative comparison at $480\times960$ ($r{=}1$) including OmniRoam.
  OmniRoam produces blurry, trajectory-inconsistent frames that diverge progressively
  from the intended path, while our model remains sharp and geometrically faithful.
  GT shown for reference; additional scenes in the supplement.}
  \label{fig:qualitative_omniroam}
\end{figure*}

\subsubsection{Zero-Shot Generalization}

Table~\ref{tab:video_realsee} evaluates on $99$
RealSee3D scenes, which provide only static RGB-D panoramas and thus have no ground-truth
videos or trajectories. OmniRoam is excluded from quantitative evaluation because, as shown in
Figure~\ref{fig:qualitative_realsee}, it produces visible wall-penetration artifacts: its
trajectory representation discards metric geometry, so NavMesh collision constraints are not
respected and the resulting frames are geometrically invalid.
Our full model achieves the best FID, best loop-closure score, and best LPIPS-Loop,
clearly outperforming Matrix-3D on all three metrics.
It also substantially reduces pose drift: relative to Matrix-3D it lowers ATE by roughly $62\%$ and RTE by $49\%$, and also improves RRE, verified via COLMAP on the generated frames. All of this is achieved with only $4$ denoising steps, showing the benefits transfer zero-shot to scenes from a different dataset.

\begin{table}[tbp]
\centering
\caption{Zero-shot evaluation on RealSee3D (no roaming-video ground truth).
Appearance/consistency metrics over $99$ scenes; pose metrics (ATE, RRE, RTE) via COLMAP over $95$ scenes.
Loop = trajectory loop-closure score (following OmniRoam~\cite{liuOmniRoamWorldWandering2026});
LPIPS-Loop = first/last-frame LPIPS. \textbf{Bold} = best.}
\label{tab:video_realsee}
\small
\setlength{\tabcolsep}{4pt}
\resizebox{\columnwidth}{!}{%
\begin{tabular}{lccccccc}
\hline
Method & Steps & FID$\downarrow$ & Loop$\uparrow$ & LPIPS-Loop$\downarrow$ & ATE$\downarrow$ & RRE\,($^\circ$)$\downarrow$ & RTE$\downarrow$ \\
\hline
Matrix-3D     & 10  & 17.29          & 2.787          & 0.1629          & 0.1176          & 0.2209          & 0.0563 \\
Ours (4-step) & \;4 & \textbf{16.36} & \textbf{3.383} & \textbf{0.1188} & \textbf{0.0448} & \textbf{0.1662} & \textbf{0.0289} \\
\hline
\end{tabular}%
}
\end{table}

\begin{figure*}[t]
  \vspace*{-1em}
  \centering
  {\setlength{\tabcolsep}{1pt}
  \renewcommand{\arraystretch}{0.95}
  \begin{tabular}{ccccccc}
    & \scriptsize Frame 0 & \scriptsize Frame 20 & \scriptsize Frame 40
    & \scriptsize Frame 60 & \scriptsize Frame 80 \\[1pt]
    \qrowGBoxFourth{OmniRoam}{panovideo/realsee_s4_synthetic_scene_06558__vp0}{omniroam}
    \qrowGBoxFourth{Matrix-3D}{panovideo/realsee_s4_synthetic_scene_06558__vp0}{matrix3d}
    \qrowGBoxFourth{Ours\\(4-step)}{panovideo/realsee_s4_synthetic_scene_06558__vp0}{ours4step}
    \noalign{\vskip 3pt\hrule\vskip 3pt}
    \qrowGBoxMiddleThree{OmniRoam}{panovideo/realsee_s4_synthetic_scene_07647__vp0}{omniroam}
    \qrowGBoxMiddleThree{Matrix-3D}{panovideo/realsee_s4_synthetic_scene_07647__vp0}{matrix3d}
    \qrowGBoxMiddleThree{Ours\\(4-step)}{panovideo/realsee_s4_synthetic_scene_07647__vp0}{ours4step}
  \end{tabular}}
  \caption{Zero-shot qualitative comparison on RealSee3D (no ground truth available).
  OmniRoam exhibits trajectory-inconsistent wall-penetration artifacts (frames pass through
  walls), which is why its metrics are not reported in Table~\ref{tab:video_realsee}. Our model generates geometrically faithful content
  and respects collision constraints with only $4$ denoising steps.
  Two representative scenes shown; each row group is one scene.}
  \label{fig:qualitative_realsee}

\medskip
\captionof{table}{Ablation of the proposed components across trajectory strides (InteriorGS, $60$ scenes).
NFE = steps $\times$ forward passes per step (CFG adds an unconditional pass).
Pose metrics (ATE, RRE, RTE) via COLMAP. \textbf{Bold} = best per trajectory stride.}
\label{tab:video_ablation}
\footnotesize
\begin{tabular}{llcccccccccc}
\hline
Method & Stride $r$ & Steps & NFE & PSNR$\uparrow$ & SSIM$\uparrow$ & LPIPS$\downarrow$ & FVD$\downarrow$ & FID$\downarrow$ & ATE$\downarrow$ & RRE\,($^\circ$)$\downarrow$ & RTE$\downarrow$ \\
\hline
Matrix-3D        & $r{=}1$ & 10 & 20 & 21.584          & 0.7977          & 0.2771          & 11.101         & 26.052          & 0.0081          & 0.0274          & 0.0021 \\
+Mixed Training  & $r{=}1$ & 10 & 20 & \textbf{23.584} & \textbf{0.8353} & \textbf{0.2220} & 8.849          & 20.209          & 0.0062          & 0.0979          & 0.0024 \\
Ours (4-step)    & $r{=}1$ & \;4 & \;4 & 23.499          & 0.8272          & 0.2330          & \textbf{7.932} & \textbf{18.249} & \textbf{0.0035} & \textbf{0.0212} & \textbf{0.0016} \\
\hline
Matrix-3D        & $r{=}2$ & 10 & 20 & 19.958          & 0.7562          & 0.3190          & 12.270         & 27.921          & 0.0312          & 0.3212          & 0.0073 \\
+Mixed Training  & $r{=}2$ & 10 & 20 & 21.921          & \textbf{0.7993} & \textbf{0.2585} & 9.500          & 21.892          & 0.0076          & 0.0619          & 0.0028 \\
Ours (4-step)    & $r{=}2$ & \;4 & \;4 & \textbf{21.925} & 0.7905          & 0.2718          & \textbf{9.228} & \textbf{20.310} & \textbf{0.0068} & \textbf{0.0285} & \textbf{0.0026} \\
\hline
Matrix-3D        & $r{=}4$ & 10 & 20 & 18.034          & 0.7064          & 0.3749          & 15.066         & 30.720          & 0.0952          & 0.1333          & 0.0147 \\
+Mixed Training  & $r{=}4$ & 10 & 20 & 20.230          & \textbf{0.7609} & \textbf{0.3040} & 10.864         & 22.144          & 0.0403          & 0.2018          & 0.0097 \\
Ours (4-step)    & $r{=}4$ & \;4 & \;4 & \textbf{20.288} & 0.7506          & 0.3194          & \textbf{10.810} & \textbf{22.133} & \textbf{0.0218} & \textbf{0.0460} & \textbf{0.0066} \\
\hline
\end{tabular}
\end{figure*}

\vspace{-0.5em}
\subsection{Feed-Forward 3DGS}
\label{sec:exp_recon}
We next evaluate the feed-forward 3D Gaussian reconstruction that lifts the
panoramic video into a real-time renderable scene.

\begin{table}[tbp]
\centering
\scriptsize
\caption{Comparison on panorama feed-forward 3DGS (trajectory stride $r{=}4$).
$N$@$N$: $N$ context views, $N$ novel views. \textbf{Bold} = best.}
\label{tab:comp_ff_3dgs}
\small
\setlength{\tabcolsep}{3.5pt}
\resizebox{\columnwidth}{!}{%
\begin{tabular}{llcccccc}
\toprule
 & & \multicolumn{2}{c}{PSNR$\uparrow$} & \multicolumn{2}{c}{SSIM$\uparrow$} & \multicolumn{2}{c}{LPIPS$\downarrow$} \\
\cmidrule(lr){3-4}\cmidrule(lr){5-6}\cmidrule(lr){7-8}
Setting & Method & Ctx. & Nov. & Ctx. & Nov. & Ctx. & Nov. \\
\midrule
\multirow{6}{*}{2@2} & DA3$^{*}$      & 17.40 & 16.45 & 0.72 & 0.70 & 0.47 & 0.53 \\
              & DA3$^{\dagger}$  & 18.80 & 17.35 & 0.74 & \underline{0.72} & 0.40 & 0.46\\
                     & HY-WM2.0$^{*}$ & 17.43 & 15.43 & 0.66 & 0.60 & 0.40 & 0.48\\
         & HY-WM2.0$^{\dagger}$  & 21.34 & 17.92 & 0.76 & 0.68 & \underline{0.29} & 0.40\\
                     & PanoWorld & 24.73 & 15.53 & \textbf{0.88} & 0.63 & \textbf{0.11} & \underline{0.37} \\
                     & Ours      & \textbf{26.75} & \textbf{19.55} & \underline{0.87} & \textbf{0.76} & 0.36 & \textbf{0.30} \\
\midrule
\multirow{6}{*}{4@4} & DA3$^{*}$      & 17.68 & 17.45 & 0.72 & 0.72 & 0.47 & 0.49 \\
              & DA3$^{\dagger}$  & 19.03 & 18.78 & 0.73 & \underline{0.73} & 0.40 & 0.42\\
                     & HY-WM2.0$^{*}$ & 19.17 & 18.11 & 0.71 & 0.68 & 0.38 & 0.42 \\
         & HY-WM2.0$^{\dagger}$  & 20.51 & \underline{19.58} & 0.74 & 0.72 & 0.34 & 0.37 \\
                     & PanoWorld & \underline{22.32} & 18.62 & \underline{0.81} & 0.69 & \textbf{0.19} & \underline{0.30} \\
                     & Ours      & \textbf{25.95} & \textbf{23.95} & \textbf{0.866} & \textbf{0.831} & \underline{0.265} & \textbf{0.299} \\
\midrule
\multirow{6}{*}{8@8} & DA3$^{*}$      & 17.95 & 17.87 & 0.73 & 0.72 & 0.46 & 0.47 \\
              & DA3$^{\dagger}$  & 19.16 & 19.10 & 0.73 & \underline{0.73} & 0.41 & 0.41\\
                     & HY-WM2.0$^{*}$ & 19.86 & 19.51 & 0.72 & 0.71 & 0.37 & 0.39\\
         & HY-WM2.0$^{\dagger}$  & 19.04 & 18.85 & 0.71 & 0.70 & 0.38 & 0.39\\
                     & PanoWorld & \underline{21.83} & \underline{19.91} & \underline{0.79} & 0.72 & \textbf{0.21} & \textbf{0.27} \\
                     & Ours      & \textbf{23.97} & \textbf{23.23} & \textbf{0.83} & \textbf{0.81} & \underline{0.31} & \underline{0.32} \\
\bottomrule
\end{tabular}%
}
\vspace{-1mm}
\\[2pt]
{\scriptsize $*$ 6-cubemap \scriptsize $^{\dagger}$ 12-cubemap}
\vspace{-3mm}
\end{table}

\begin{figure*}[t]
  \centering
  \setlength{\tabcolsep}{1pt}
  \renewcommand{\arraystretch}{0.95}
  \begin{tabular}{cccccc}
    \qrowGBoxfddgs{Ground Truth}{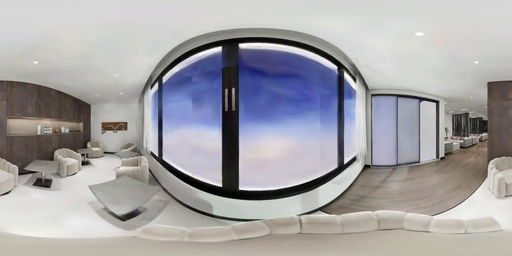}
    \qrowGBoxfddgs{HY-WM2.0}{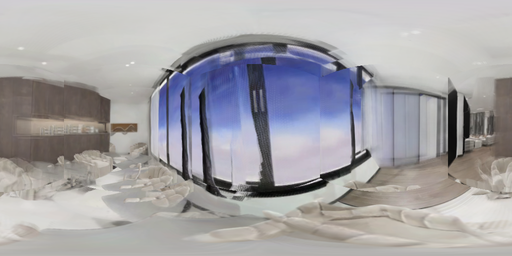}
    \qrowGBoxfddgs{PanoWorld}{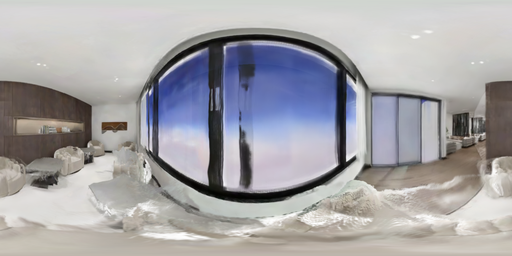}
    \qrowGBoxfddgs{Ours}{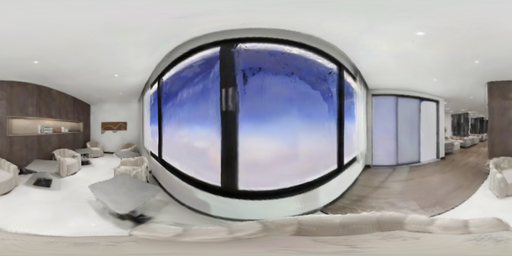}
  \end{tabular}
  \caption{Qualitative comparison of novel view synthesis on the InteriorGS dataset (60 scenes). We evaluate all methods under the 4@4 setting.}
  \label{fig:qualitative_feedforwardgs}
\end{figure*}

\paragraph{Quantitative Comparison.} 
Table~\ref{tab:comp_ff_3dgs} compares our method against DA3, HY-WM2.0, and PanoWorld~\cite{jiaPanoWorldGenerativeSpatial2026} on the $4\times$-speed test set, where we uniformly sample $N$ input (context) views and $N$ held-out novel views per scene ($N\!\in\!\{2,4,8\}$, denoted $N$@$N$). To isolate the contribution of the reconstruction decoder, we use real captured videos rather than trajectory-controllable generated videos, eliminating the extra variable introduced by generative artifacts. Two observations can be made. \textbf{First}, our feed-forward 3DGS decoder consistently outperforms all baselines on both context-view reconstruction and novel-view synthesis across all three settings. For example, at $4$@$4$ our method achieves $25.95$/$23.95$ PSNR on context/novel views, surpassing the best baseline PanoWorld by $3.63$/$5.33$ dB, and similarly leads in SSIM while remaining competitive in LPIPS. This indicates that our decoder recovers accurate Gaussians from geometrically consistent observations even under sparse, wide-baseline panoramic coverage. We further provide qualitative comparisons in Fig.~\ref{fig:qualitative_feedforwardgs}, where our method produces sharper textures and more faithful geometry than the baselines. \textbf{Second}, we observe that the rendering quality on context views gradually degrades as the number of reconstructed images increases (e.g., PSNR drops from $26.75$ at $2$@$2$ to $25.95$ at $4$@$4$ and further to $23.97$ at $8$@$8$). This is because, with more images involved in reconstruction, even slight misalignment across views accumulates and is amplified during voxel fusion, which in turn noticeably degrades the final rendering quality.

\subsection{Ablation Study}

Table~\ref{tab:video_ablation} isolates the contribution of each component across all three traversal speeds and reports NFE, accounting for both step count and the unconditional branch of classifier-free guidance. Long--short mixed training contributes most of the quality gain ($+2.0$\,dB PSNR, $-5.8$ FID at $s{=}1\times$), and the improvement grows with traversal distance. Shortcut self-consistency training then removes the unconditional branch (CFG-free) and cuts the step count from $10$ to $4$, reducing NFE by $5\times$ \emph{without} a loss in fidelity; it attains the best FVD, FID, and pose accuracy at all speeds. This combination is what makes deployment on a single consumer GPU feasible: the $4$-step CFG-free sampler reduces diffusion cost from ${\sim}316$\,s to $63$\,s, bringing the complete per-scene pipeline to ${\sim}147$\,s on a single NVIDIA RTX~4090 (24\,GB) with no per-scene optimization.

\paragraph{Ablation of decoder components.}
Table~\ref{tab:3dgs_ablation} isolates the contribution of each attention mechanism in the
Gaussian decoder under the $4$@$4$ setting. Removing the KNN attention blocks
(w/o KNN Attn) drops novel-view PSNR by $0.80$\,dB and increases LPIPS by $0.022$,
confirming that local 3D neighborhood reasoning is critical for resolving geometric
ambiguities in the voxel tokens. Removing the multi-view attention blocks (w/o MV Attn)
causes a smaller but consistent degradation across all metrics, indicating that cross-view
feature exchange helps disambiguate occluded regions and improves multi-view consistency.
The full model combining both mechanisms achieves the best performance on all metrics.
\begin{table}[ht]
\centering
\caption{Ablation of the 3D Gaussian decoder components ($4$@$4$ setting, $4\times$ speed test set). \textbf{Bold} = best.}
\label{tab:3dgs_ablation}
\small
\setlength{\tabcolsep}{3.5pt}
\begin{tabular}{lcccccc}
\toprule
 & \multicolumn{2}{c}{PSNR$\uparrow$} & \multicolumn{2}{c}{SSIM$\uparrow$} & \multicolumn{2}{c}{LPIPS$\downarrow$} \\
\cmidrule(lr){2-3}\cmidrule(lr){4-5}\cmidrule(lr){6-7}
Method & Ctx. & Nov. & Ctx. & Nov. & Ctx. & Nov. \\
\midrule
w/o voxel fusion & 24.34 & 23.01 & 0.837 & 0.810 & 0.301 & 0.336 \\
w/o KNN Attn & 25.00 & 23.15 & 0.849 & 0.818 & 0.293 & 0.321 \\
w/o MV Attn  & 25.66 & 23.45 & 0.851 & 0.827 & 0.273 & 0.301 \\
1 gaussian per voxel & 25.12 & 23.13 & 0.847 & 0.822 & 0.273 & 0.317 \\
Full Model   & \textbf{25.95} & \textbf{23.95} & \textbf{0.866} & \textbf{0.831} & \textbf{0.265} & \textbf{0.299} \\
\bottomrule
\end{tabular}
\end{table}


\FloatBarrier
\subsection{Joint Analysis of Generation and 3D Reconstruction}
\label{sec:joint_analysis}

We directly compare COLMAP and our feed-forward VGGT estimator as pose recovery tools,
on both ground-truth frames and our generated frames across all trajectory strides.
Table~\ref{tab:recon_gt} evaluates on GT frames (upper bound) and
Table~\ref{tab:recon_gen} evaluates on our generated frames (pipeline output).
Table~\ref{tab:recon_render} reports the end-to-end 3DGS rendering quality from
the generated videos.

On GT frames, COLMAP achieves near-zero ATE ($\leq 0.003$) while VGGT has a systematic
noise floor around $0.023$--$0.030$, so COLMAP is the more precise absolute estimator. On
generated frames, COLMAP reveals the true trajectory drift as traversal distance grows
($r{=}4$: ATE $0.022$), whereas VGGT remains more stable (ATE $0.035$) because it
relies on local feature matching that is partially robust to global drift. The gap between
GT and generated frames is small for VGGT at all trajectory strides (e.g.\ $0.029$ vs.\ $0.030$ at
$r{=}1$), which shows that our generation quality preserves enough texture and
structure for the reconstruction module to recover locally consistent metric poses for 3DGS.
The 3DGS decoder then achieves $18.6$--$19.7$\,dB novel-view PSNR across all trajectory strides
(Table~\ref{tab:recon_render}), giving a complete and usable reconstruction pipeline.

\begin{table}[tbp]
\centering
\caption{Pose estimation on \emph{generated frames} (Ours, 4-step): COLMAP vs.\ our
feed-forward VGGT estimator. COLMAP exposes trajectory drift; VGGT measures pose
recovery quality for downstream 3DGS.}
\label{tab:recon_gen}
\small
\setlength{\tabcolsep}{4pt}
\begin{tabular}{llccc}
\hline
Estimator & Stride $r$ & ATE$\downarrow$ & RRE\,($^\circ$)$\downarrow$ & RTE$\downarrow$ \\
\hline
COLMAP          & $r{=}1$ & \textbf{0.0035} & \textbf{0.0212} & \textbf{0.0016} \\
VGGT (ours)     & $r{=}1$ & 0.0290 & 0.0542 & 0.0282 \\
\hline
COLMAP          & $r{=}2$ & \textbf{0.0068} & \textbf{0.0285} & \textbf{0.0026} \\
VGGT (ours)     & $r{=}2$ & 0.0236 & 0.0531 & 0.0269 \\
\hline
COLMAP          & $r{=}4$ & \textbf{0.0218} & \textbf{0.0460} & \textbf{0.0066} \\
VGGT (ours)     & $r{=}4$ & 0.0346 & 0.0526 & 0.0357 \\
\hline
\end{tabular}
\end{table}

\begin{table}[tbp]
\centering
\caption{Pose estimation on \emph{ground-truth frames} at trajectory stride $r{=}4$: COLMAP vs.\ our feed-forward VGGT estimator. Lower values indicate higher precision.}
\label{tab:recon_gt}
\small
\setlength{\tabcolsep}{5pt}
\begin{tabular}{lcccc}
\toprule
Estimator   & ATE$\downarrow$ & ATE-RMSE$\downarrow$ & RPE$_t$$\downarrow$ & RPE$_r$($^\circ$)$\downarrow$ \\
\midrule
COLMAP      & \textbf{0.0042} & \textbf{0.0046} & \textbf{0.0019} & \textbf{0.03} \\
PanoVGGT    & 0.0070          & 0.0083          & 0.0045          & 0.51          \\
VGGT (ours) & 0.0170          & 0.0218          & 0.0148          & 0.04          \\
\bottomrule
\end{tabular}
\end{table}

\begin{table}[tbp]
\centering
\caption{3DGS rendering quality on generated frames (Ours, 4-step).
\emph{Inp.}\ = input-view reconstruction; \emph{Nov.}\ = novel-view synthesis.}
\label{tab:recon_render}
\small
\setlength{\tabcolsep}{4pt}
\begin{tabular}{lcccccc}
\hline
Stride $r$ & \multicolumn{2}{c}{PSNR$\uparrow$} & \multicolumn{2}{c}{SSIM$\uparrow$} & \multicolumn{2}{c}{LPIPS$\downarrow$} \\
      & Inp. & Nov. & Inp. & Nov. & Inp. & Nov. \\
\hline
$r{=}1$ & 21.08 & 19.45 & 0.73 & 0.67 & 0.22 & 0.26 \\
$r{=}2$ & 22.06 & 19.68 & 0.76 & 0.67 & 0.20 & 0.26 \\
$r{=}4$ & 22.56 & 18.64 & 0.78 & 0.64 & 0.20 & 0.31 \\
\hline
\end{tabular}
\end{table}

\FloatBarrier
\section{Discussion and Limitations}

Genie Sim PanoWorld demonstrates that a single ERP panorama is sufficient to produce a freely
roamable 3D Gaussian scene entirely feed-forward, without any per-scene optimization.
The key enabler is the two-stage design: the trajectory-controllable video generator
converts a single view into wide-baseline, geometrically consistent multi-view evidence,
and the metric $\mathrm{SE}(3)$ trajectory ties both stages into one coherent frame.
Long--short mixed training and self-consistency training for shortcut models make this practical on commodity
hardware, substantially reducing NFE relative to a CFG-guided baseline while maintaining
or improving all reported quality metrics.

Several limitations remain.
First, the pipeline assumes a static scene: moving objects such as people or
furniture are not modeled, and their presence can introduce temporal inconsistencies in
the generated video.
Second, NavMesh construction depends on the quality of the monocular depth estimate;
specular reflections, transparent surfaces, and large textureless planes can produce
depth errors that lead to suboptimal trajectories or residual wall-penetration artifacts.
Third, the geometry-warped conditioning becomes increasingly unreliable beyond the
trajectory strides tested ($r{=}4$), where the fraction of disoccluded content
can exceed the model's inpainting capacity.
Finally, the current training distribution is dominated by indoor environments, and
performance on outdoor or large-scale scenes has not been evaluated.

Future work can address these limitations in several ways.
Incorporating a video-object segmentation prior to mask dynamic content during training
and inference would decouple scene geometry from transient objects.
Using a more robust depth-completion approach or fine-tuning the depth backbone on panoramic
data with known metric scale would improve NavMesh quality.
Extending the shortcut-model framework to longer-context video diffusion models
and larger-scale would push the navigable range further.
Finally, adapting the 3D Gaussian decoder to outdoor and mixed-scale environments is a
natural step toward broader real-world deployment.

\section{Conclusion}

We presented Genie Sim PanoWorld, a feed-forward pipeline that turns a single $360^\circ$ ERP
panorama into a freely roamable 3D Gaussian scene without per-scene optimization or
multi-view capture.
The core insight is 
to bridge the single-view input and the 3D output with an explicit,
trajectory-controllable panoramic video: a NavMesh-planned $\mathrm{SE}(3)$ path is
injected via dense geometry-warped conditioning into a latent video diffusion model
trained with long--short mixed supervision and a self-consistency objective for shortcut models,
producing high-fidelity panoramic video in four CFG-free steps; a feed-forward
panoramic reconstructor then lifts the result into a real-time renderable 3DGS scene.
Experiments show that Genie Sim PanoWorld consistently outperforms prior baselines in
both video fidelity and trajectory accuracy while running the complete pipeline
efficiently on a single consumer-grade GPU.
We hope Genie Sim PanoWorld serves as a practical foundation for panorama-driven scene
understanding, interactive virtual tours, and embodied AI navigation.

\bibliographystyle{ieeenat_fullname}
\bibliography{reference}

\clearpage
\end{document}